# CLD-shaped Brushstrokes in Non-Photorealistic Rendering

A.C. Sparavigna [1] and R. Marazzato [2]

[1] Dipartimento di Fisica, Politecnico di Torino, Italy

[2] Dipartimento di Automatica ed Informatica, Politecnico di Torino, Italy
amelia.sparavigna@polito.it, roberto.marazzato@polito.it

**Abstract:** *Rendering techniques based on a random grid can be improved by adapting brushstrokes to the shape of different areas of the original picture. In this paper, the concept of Coherence Length Diagram is applied to determine the adaptive brushstrokes, in order to simulate an impressionist painting. Some examples are provided to instance the proposed algorithm.*

Coherence Length Diagram.

## 1 Introduction

Non-photorealistic rendering is focused on developing algorithms for generating or processing images that embody qualities such as the emphasis of certain features, the suppression of details or the use of stylization to suggest specific feelings in the observer. In fact, a wide variety of expressive styles is already existing [1, 2, 3, 4]: among them we can find non-photorealistic rendering styles, inspired by artistic methods such as painting and drawing, useful for digital art. Many rendering techniques simulate the painter's medium, with methods emulating the diffusion of colours through different canvas and paper media to produce watercolour or oil painting effect. Our previous paper [5] describes some non photorealistic rendering algorithms which reproduce different features of impressionist painting. The common framework of such methods consists of two steps: creating a random grid of pixel positions, where to place brushstrokes, and determining the shape and the color of each brushstroke as a function of the color of nearby pixels and of some other parameters. In our previous work the shape was limited to be a circle, a rectangle or a square. The aim of this paper is to shape brushstrokes across the image according to local orientations a painter could manually follow in order to reproduce the texture of the subject. For instance, the upper left area of the picture in fig.1 should be textured with a set of straight vertical strokes, while the lower right area should correspond to angular strokes.

The behaviour of CLD diagrams (see [6],[7],[8]) approximates this requirement, as each pixel position can be associated to a local diagram which reflects both the orientation and the length of the image texture for that point. Such shapes only need to be rescaled and filled with the appropriate RGB color, as in previous algorithms.

## 2 CPB Non Photorealistic Rendering

The starting point for CLD shaping of brushstrokes is the simpler Circular Paintbrush (CPB) algorithm, described in [5]: the RGB source image of $N_x \times N_y$ pixels is reppresented by the three-

channel brightness function
$$b : I \to B,$$
$$I = [1, N_x] \times [1, N_y] \subset \mathbf{N}^2,$$
$$B = [0, 255]^3 \subset \mathbf{N}^3.$$

For each step $k$ in the sequence $k \in [1, \cdots N] \subset \mathbf{N}$, a regularly spaced grid of pixel positions $G_k$ is generated as

$$G_k = \{(p_i^k, p_j^k) \in I : p_i^k = i s_k, p_j^k = j s_k\};$$
$$i, j, s_k \in \mathbf{N}; p_i^k \leq N_x; p_j^k \leq N_y$$

where $s_k \in \mathbf{N}$ is a random value with a suitable distribution.

The random grid $G_k'$ is then computed from $G_k$ by adding to each position a displacement

$$(\delta_{ij}^{(k,x)}, \delta_{ij}^{(k,y)}) \in [-\Delta, +\Delta]^2 \subset \mathbf{Z}^2.$$

so obtaining

$$G_k' = \{(p_i^k + \delta_{ij}^{(k,x)}, p_j^k + \delta_{ij}^{(k,y)}) \cap I : (p_i^k, p_j^k) \in G_k\}.$$

For each pixel position $(p_i'^k, p_j'^k) \in G_k'$ in the random grid, a circle of fixed radius $r$ is generated and filled with the solid color given by the color of the corresponding regularly spaced grid pixel $b(p_i^k, p_j^k)$ in the original image (displaced fill). An alternate solution is using the original color of the same position $b(p_i'^k, p_j'^k)$ (non displaced fill).

The previously discussed procedure is iterated several times according to the final desired covering of the canvas. In fact, CPB circles generated at each iterative step can partially or completely overlap. On the other side, some pixels of the output image might not be covered by any circle; in this case, a default color value is assigned, for instance white, the color of the corresponding pixel in the original image or an intermediate value.

## 3 The Coherence Length Diagram (CLD)

This section describes a new paint-brush based on CLD image processing. This image analysis method has been discussed in some previous works [7],[8]. Its interesting feature resides in the diagram it generates, which can be regarded as a synthetic way to represent the local texture of the image. Instead of considering the full RGB picture under conversion, the corresponding grayscale image is computed, as CLD is defined on a scalar brightness matrix. So, let us consider

$$\tilde{b}(x, y) = \frac{1}{3} \sum_{i=1}^{3} b_i(x, y)$$

where $b(x, y) = (b_1(x, y), b_2(x, y), b_3(x, y))$ and the subscripts 1, 2, 3 correspond to the channels R, G, B. First we compute the average brightness over the whole image

$$M_0 = \frac{1}{N_x N_y} \sum_{i=1}^{N_x} \sum_{j=1}^{N_y} \tilde{b}(i, j),$$

then consider a set of N=32 directions

$$\theta_i = i \frac{2\pi}{32}; i = 1, \ldots, 32.$$

For each point $(x, y) \in I$, we can define the local first order moment

$$M_{0,\ell}^{i}(x,y) = \frac{1}{\ell}\sum_{r=0}^{\ell}\tilde{b}(x+\lfloor r\cos\theta_i\rfloor, y+\lfloor r\sin\theta_i\rfloor),$$

which is a function of the distance $\ell$. The local CLD, for a given threshold $\tau$, represents the minimum summing distance $\ell$ such that the corresponding first order moment differs from $M_0$ by less than $\tau$:

$$l_{0,i}^{\tau}(x,y) = \min\left\{\ell : \frac{|M_{0,\ell}^{i}(x,y) - M_0|}{M_0} \leq \tau\right\}. \tag{1}$$

This quantity is not defined for all directions of all point: sometimes the edge of the image is reached without entering the desired neighborhood of $M_0$. The appearance of a local CLD is represented in fig.2. To see other local behaviour of Coherence Length Diagram, Ref.[6] is quite suitable, as the more general discussion proposed in [7].

## 4 Shaping and filling brushstrokes

By connecting all vertices of the CLD in ascending order, a star shaped polygon [9] $P_\tau(x,y)$ is generated, in which the point $(x,y)$ belongs to the kernel. With an appropriate scale factor $\alpha$, such surface can be considered as the brushstroke at the point where the CLD is computed. The above described methods can be used together to obtain an output image containing "well shaped" and "well oriented" brushstrokes, as required at the beginning of this discussion. Instead of generating a set of circles, like in the CPB algorithm,

$$b \to \{C[p_i', p_j', r, b(p_i, p_j)]\} \tag{2}$$

$$(p_i, p_j) \in \bigcup_k G_k, (p_i', p_j') \in \bigcup_k G_k',$$

a set of CLDs

$$b \to \{\alpha P_\tau[p_i', p_j', b(p_i, p_j)]\} \tag{3}$$

$$(p_i, p_j) \in \bigcup_k G_k, (p_i', p_j') \in \bigcup_k G_k',$$

is considered. Clearly, in equations (2), (3) $C[x,y,r,c]$ represents a circle centered in $(x,y)$, having a radius $r$ and filled with a color $c$, while $P_\tau[x,y,c]$ represents a CLD polygon centered in $O=(x,y)$, obtained with a threshold $\tau$ and filled with a color $c$. The parameter $\alpha$ is the rescaling factor which allows to maintain the shape of te generated polygon and enlarge or shrink it. In a practical case, the average size

$$\overline{L} = \frac{\alpha}{32}\sum_{i=0}^{31} l_{0,i}^{\tau}(x,y)$$

of the brushstroke is set, and $\alpha$ can be consequently computed.

Again, the actual appearance of the output image depends on the sequence of steps $k$: the latest shapes override the older ones. As stated before, (2) and (3) represent a case with color displacement. The alternate choice would be implemented by filling the generated shapes with the color $b(p_i', p_j')$. The filling algorithm can be described as the sequence of the following steps:

1. Consider the vertex #0 (pivot point) and the segment $s_1$ connecting the centroid to the

vertex #1 (pivot segment). Divide $s_1$ into $N_1$ equal parts, each corresponding to one pixel, so btaining the set of points $\{P_{1,0}, \cdots P_{1,N_1}\}$, where $P_{1,0}=O$.

2. Connect each of such points $P_{1,i}$ to the vertex #0 with the corresponding segment $s_{1,i}$ and draw it - see fig. (3).

3. Repeat the previous steps swithcing the pivot point and segment form $(\#0,\#1)$ to $(\#1,\#2), \cdots, (\#30,\#31), (\#31,\#0)$.

As pixels of different segments overlap in the area around $O$, the step (2) can be improved by only drawing the part of the segments $s_{1,i}$ starting at a distance $d_0$ for $i \geq 1$, as in fig. (4). In order to estimate the appropriate value of $d_0$, we must recall that $\angle(s_i, s_{i+1}) = \frac{\pi}{16}$. After having traced $s_{1,0} = s_0$, drawing can start where two subsequent segments are two pixels away from each other. We thus obtain

$$2 = d_0 \left(\frac{\pi}{16}\right) \Rightarrow d_0 \cong 10.$$

Lower values can be used for $d_0$ as a conservative measure.

## 5 Examples

Fig.(5) compares the original image on the left with the rendering obtained with CLD paintbrush in the middle and a CPB on the right. The CLD setting had a threshold $\tau = 0.2$, a size $\overline{L} = 2px$, and used the algorithm without displacement. Another example is the tiger shown in Fig.(6), with the original image on the left and its rendering on the right. The CLD setting had again a threshold $\tau = 0.2$ and size $\overline{L} = 2px$, without displacement. Looking at Fig.7, please note how the whiskers of the tiger in the water are completely rendered.

The last example, Fig.8 is concerning the role of displacement fill on the final rendering. The image shows four CLDPB transformations for size $\overline{L} = 4px$ and $\overline{L} = 8px$, built without and with the displacement filling. Note how the rendering turns out to be different, in particular in depicting the edges of objects. The described CLD paintbrush technique has be organized in a single tool, with the procedures previously discussed in [5]. The package will be proposed in a shortcoming paper devoted to this tool illustration.

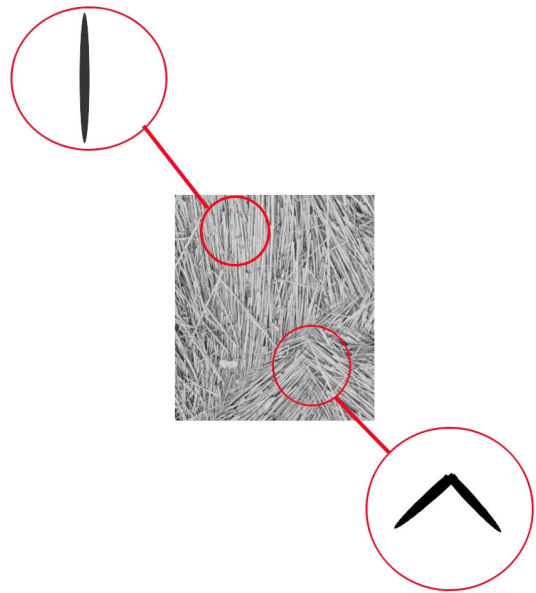

Figure 1: Brushstroke shapes for specific areas of an image

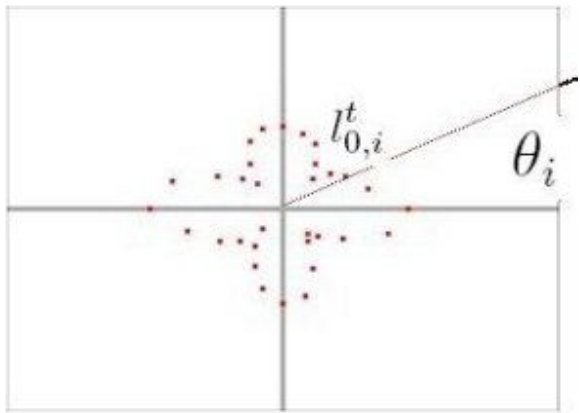

Figure 2: Example of local CLD at (x,y)

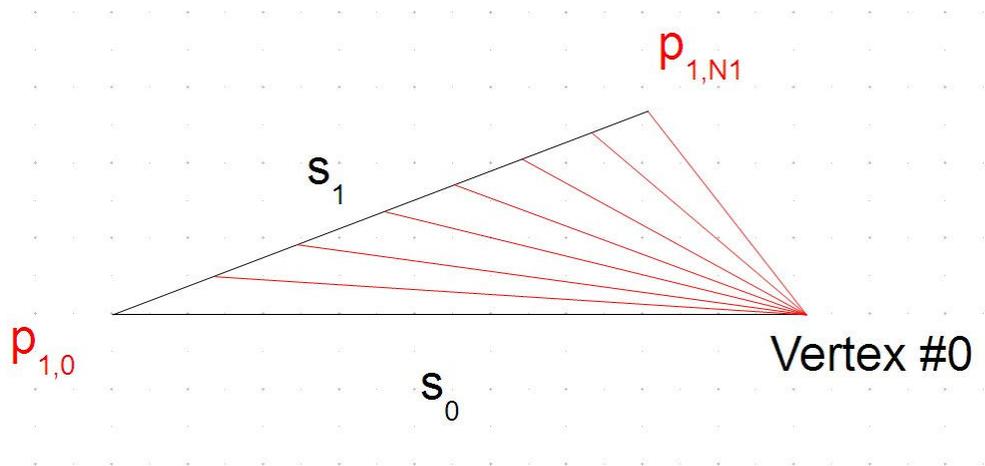

Figure 3: Fill algorithm

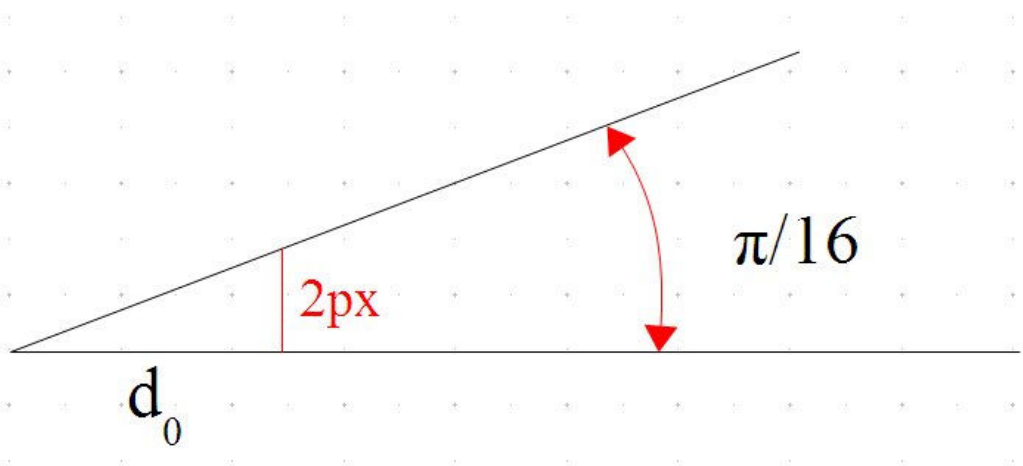

Figure 4: Minimum tracing distance

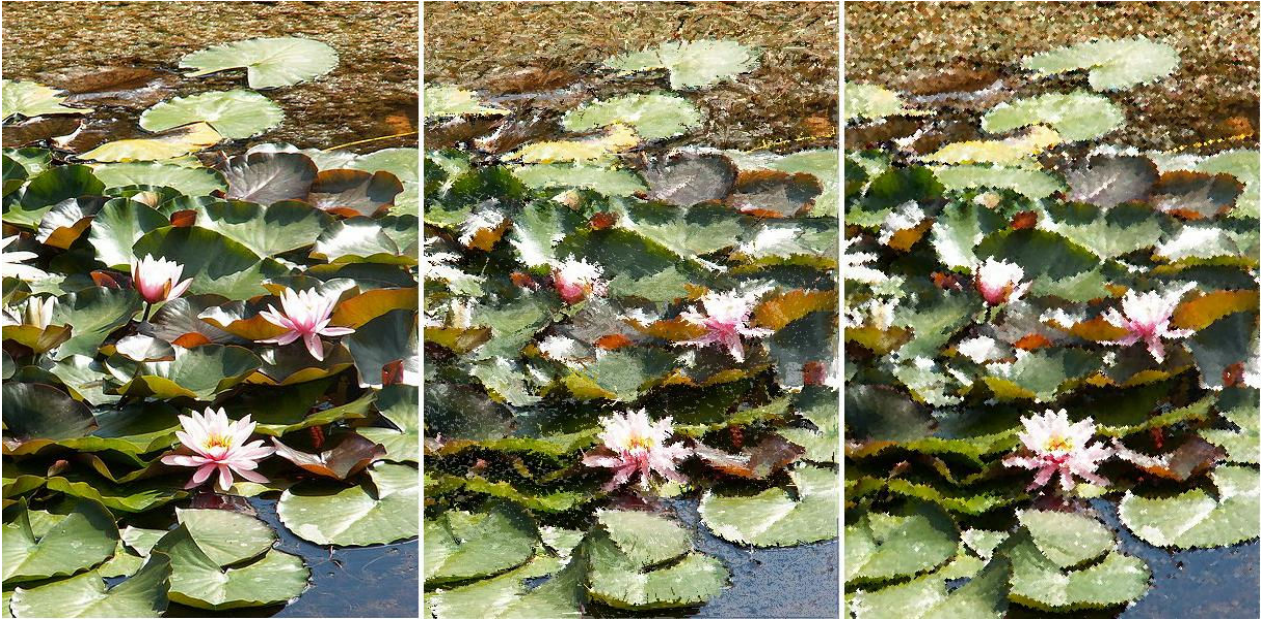

Figure 5: Original image on the left. In the middle, the CLDPB transform and on the right a CPB trasform. The original image has title "Nymphaea alba", by Opuntia, from Wikipedia.

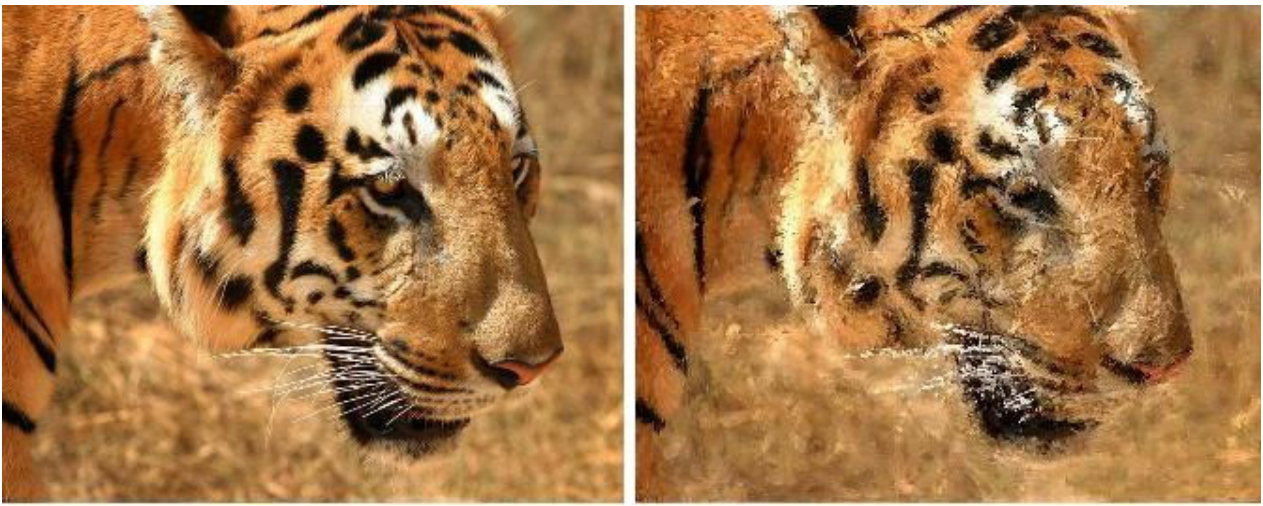

Figure 6: Original image (left) compared with its CLDPB transform (right). The original image has title "Tiger in South India", by Steve Evans, from Wikipedia.

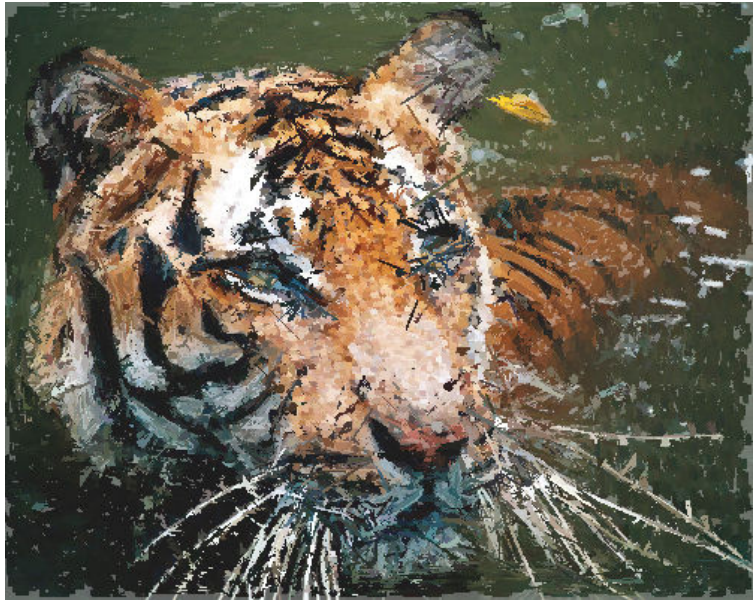

Figure 7: CLDPB transform of the "Tiger in the water", by B_cool, from SIN, from Wikipedia.

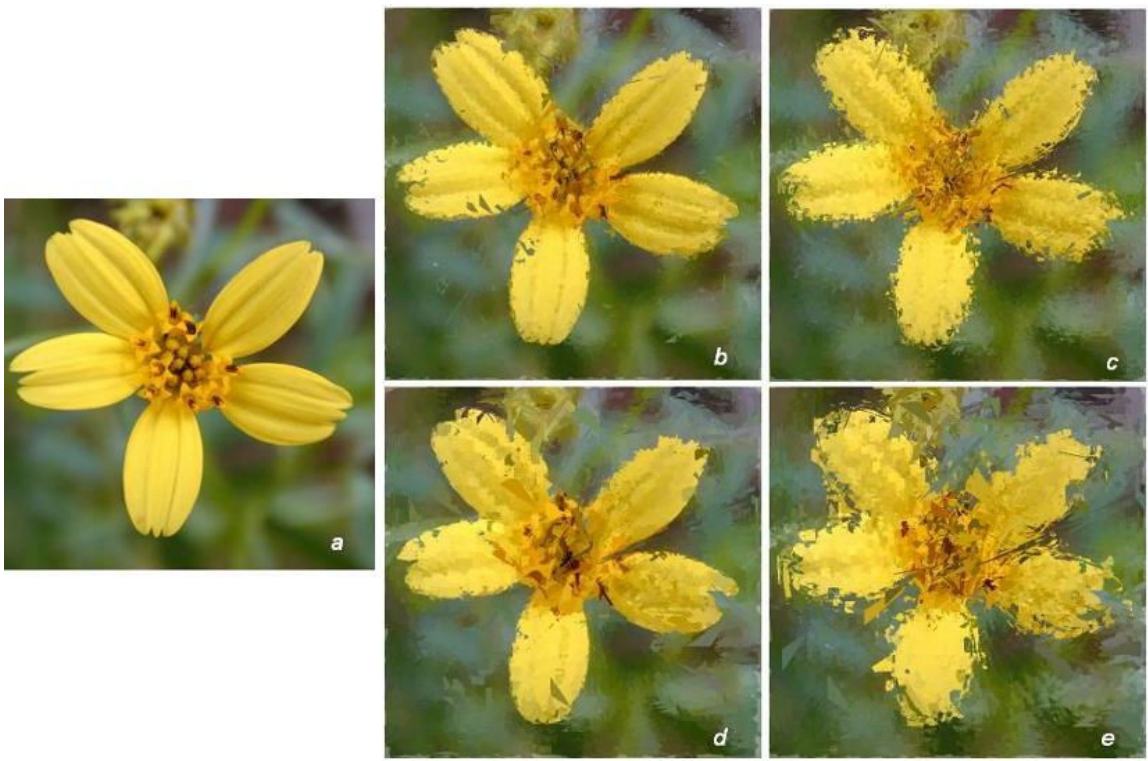

Figure 8: Image (a) shows the original image by Eric Guinther, from Wikipedia. The other images are CLDPB transformations with threshold $\tau = 0.2$. (b) and (c) were obtained with size $\overline{L} = 4\,px$, without and with displacement, respectively. (d) and (e) with $\overline{L} = 4\,px$, without and with displacement. Note the different rendering of edges.